\documentclass[sigconf]{acmart}
\pdfoutput=1
\usepackage{booktabs} 

\usepackage{url}
\usepackage{amssymb}
\usepackage{graphicx}
\usepackage{amsmath}
\usepackage{amsthm}
\usepackage{algorithm}
\usepackage{algorithmic}
\usepackage{subfig}

\copyrightyear{2020} 
\acmYear{2020} 
\setcopyright{acmlicensed}\acmConference[GECCO '20]{Genetic and Evolutionary Computation Conference}{July 8--12, 2020}{Cancún, Mexico}
\acmBooktitle{Genetic and Evolutionary Computation Conference (GECCO '20), July 8--12, 2020, Cancún, Mexico}
\acmPrice{15.00}
\acmDOI{10.1145/3377930.3390197}
\acmISBN{978-1-4503-7128-5/20/07}

\begin{document}

\title{Safe Crossover of Neural Networks Through Neuron Alignment}

\author{Thomas Uriot}
\affiliation{%
  \institution{European Space Agency}
  \city{Noordwijk} 
  \state{The Netherlands} 
}
\email{thomas.uriot@esa.int}

\author{Dario Izzo}
\affiliation{%
  \institution{European Space Agency}
  \city{Noordwijk} 
  \state{The Netherlands} 
}
\email{dario.izzo@esa.int}

\begin{abstract}
One of the main and largely unexplored challenges in evolving the weights of neural networks using genetic algorithms is to find a sensible crossover operation between parent networks. Indeed, naive crossover leads to functionally damaged offspring that do not retain information from the parents. This is because neural networks are invariant to permutations of neurons, giving rise to multiple ways of representing the same solution. This is often referred to as the competing conventions problem. In this paper, we propose a two-step \textit{safe crossover} (SC) operator. First, the neurons of the parents are functionally aligned by computing how well they correlate, and only then are the parents recombined. We compare two ways of measuring relationships between neurons: Pairwise Correlation (PwC) and Canonical Correlation Analysis (CCA). We test our safe crossover operators (SC-PwC and SC-CCA) on MNIST and CIFAR-10 by performing arithmetic crossover on the weights of feed-forward neural network pairs. We show that it effectively transmits information from parents to offspring and significantly improves upon naive crossover. Our method is computationally fast, can serve as a way to explore the fitness landscape more efficiently and makes safe crossover a potentially promising operator in future neuroevolution research and applications. 
\end{abstract}

\ccsdesc[300]{Computing methodologies~Continuous space search}
\ccsdesc[300]{Computing methodologies~Genetic algorithms}
\ccsdesc[500]{Computing methodologies~Neural networks}

\keywords{Canonical correlation analysis, competing conventions, crossover, genetic algorithms, neuroevolution}

\maketitle

\section{Introduction}

Neuroevolution~\cite{stanley2019designing} (NE) is concerned with the evolution of neural network topology, weights and learning rules. Historically, the first attempts at evolving neural networks made use of a particular type of neuroevolution technique called genetic algorithms~\cite{holland1975adaptation,goldberg1988genetic} (GAs). Genetic algorithms are population-based and aimed at loosely replicating the process of natural selection by having bio-inspired operators such as mutation, crossover (recombination) and selection, acting on the population individuals. In particular, the crossover operator picks a number of high performing parents to create offspring which share some common genotype (e.g. topology, weights, hyperparameters) with the parents. The use of crossover in neuroevolution assumes that one can find a sensible way to mix information from the parent networks, in order to create a promising offspring. However, we cannot naively recombine neural networks in a structural way by simply matching their topologies. This is due to the fact that neural networks can encode the same solution while having different representations (e.g. permuted neurons) or even different topologies. The aforementioned problem of one phenotype having multiple genotypes is often referred to as the \textit{competing conventions problem}, or the \textit{permutation problem}~\cite{montana1989training,schaffer1992combinations,radcliffe1993genetic}, where a convention corresponds to a particular network representation. If the competing conventions problem is not addressed, crossover essentially becomes an inefficient and even detrimental operator which produces dysfunctional offspring. In addition, the competing conventions problem means that a unimodal fitness landscape to a local gradient-based algorithm can become multimodal to GAs, with each mode corresponding to a unique convention~\cite{schaffer1992combinations}. This is due to the fact that local optimization such as gradient descent only looks at the direct neighborhood of the current solution point, while global optimization algorithms such as GAs explore the whole search space. Thus, the resulting search space for GAs is bigger by a factor of $(n!)^D$ for a network with $D$ hidden layers of size $n$, than it really ought to be~\cite{thierens1996non}. The challenge to transmit information efficiently from parents to offspring, is to solve the competing conventions problem. This is done by finding a non-redundant~\cite{radcliffe1993genetic,thierens1996non} (i.e. unique) representation of the set of functionally equivalent neural networks, by permuting neurons in order to functionally align the parent networks~\cite{li2015convergent}.

While a lot of early work had relative success in applying GAs with naive crossover to evolve small neural networks on simple reinforcement learning (RL) tasks~\cite{whitley1989genitor,wieland1991evolving,moriarty1996efficient,gomez1999solving,gomez2008accelerated}, two main approaches to evolve neural networks arose. The first one concluded that crossover is harming the evolutionary process and thus that classical GAs~\cite{holland1975adaptation,goldberg1988genetic} are not well suited to evolve neural networks~\cite{yao1998towards,angeline1994evolutionary}. This neuroevolution paradigm is called evolutionary programming, where a population of neural networks is evolved by solely relying on mutations and selection. 
On the other hand, the second approach continued using crossover and attempted to tackle the competing conventions problem~\cite{thierens1996non,montana1989training,stanley2002evolving}.~\citet{thierens1996non}, addresses the representation redundancy by defining a mapping for which any networks from the same symmetry group is transformed to a unique representation. The mapping consists in permuting the neurons by ranking them according to their bias values in ascending order. This mapping, however, does not take into account neuron functionalities, which are key for the offspring to retain information from the parents. More related to our work, ~\citet{montana1989training}, attempt to crossover networks by matching neurons functionalities by comparing their responses to inputs. Another approach, the NeuroEvolution of Augmenting Topologies (NEAT) algorithm~\cite{stanley2002evolving} aims at incrementally growing networks from minimal structure. In order to perform crossover between different networks, the authors match neuron functionalities by using markers which keep track of the neuron origins (parents). Only neurons sharing the same origins are allowed to be recombined together. Recently, \citet{gangwani2017policy} worked around the competing conventions problem by introducing crossover directly in state space (in an RL context), using imitation learning to learn a child policy which combines the strength of the two parent policies.

In our work, we use the neuron representation introduced in \citet{li2015convergent}, which defines a neuron by its activation values over a batch of inputs (see Section 3.1). With this representation, we consider two ways of characterizing the relationships between neurons: Canonical Correlation Analysis~\cite{hotelling1992relations,uurtio2018tutorial,raghu2017svcca,morcos2018insights} and simple pairwise correlation~\cite{li2015convergent}. We then propose two safe crossover operators: SC-CCA and SC-PwC, depending on whether the neurons are mapped using CCA or pairwise correlation, respectively. Our safe crossovers proceed in two steps: first the neurons of the parent networks are functionally aligned, and only then are the parents allowed to be recombined. By functional alignment, we mean that neurons located at the same position in the parent networks should have learned the same internal representation of the data, which is measured by how well they correlate (using CCA or pairwise correlation). However, it is not always possible to align neurons as a particular network may have learned unique internal representations that other networks have not~\cite{li2015convergent}. Furthermore, many neurons simply capture noise in the data and do not have clear counterparts. 

To the best of our knowledge, this is the first attempt at defining a safe crossover operator acting directly in parameter space. Interestingly, this process is analogous to meiotic division found in nature, where homologous chromosomes are first aligned before being recombined ~\cite{finsterbusch2016alignment,buonomo2000disjunction}, taking the parallel between GAs and natural evolution a step further. To summarise, our contributions are as follows:
\begin{itemize}
    \item In Section 2, we review some of the relevant literature, in the context of our work. In addition, we argue that it is an important research question with the potential to further enhance the use of GAs in neuroevolution applied to RL tasks, as well as improving direct exploration in parameter space. 
    \item In Section 3 and 4, we build on the work from ~\cite{raghu2017svcca,morcos2018insights,li2015convergent} and use Canonical Correlation Analysis as well as pairwise cross-correlation in order to find mappings between neurons of neural network pairs. This allows us to permute neurons in order to functionally align the networks. 
    \item In Section 5, we perform a full (i.e. zero-point) arithmetic crossover on each pair of aligned networks. We compare our two safe crossover operators on MNIST and CIFAR-10 and show that the produced offspring preserve functions from the parents, as opposed to naive crossover. Furthermore, we show that the offspring produced via safe crossover outperform the parents by achieving a better validation loss. 
    \item Finally, in Section 6, we reflect on our findings and discuss further improvements as well as potential applications of safe crossover.
\end{itemize}

\section{Motivation}

\subsection{Evolutionary Algorithms in Reinforcement Learning}

Recently, with the increase of computational power available, and due to their highly parallelizable nature, evolutionary algorithms were successfully applied to modern reinforcement learning benchmarks~\cite{salimans2017evolution,such2017deep,mania2018simple,lehman2018safe,koutnik2014evolving,gangwani2017policy}, rivaling gradient-based methods such as Trust Region Policy Optimization~\cite{schulman2015trust}, Deep-Q learning~\cite{mnih2015human} and A3C~\cite{mnih2016asynchronous}. Evolutionary Strategies~\cite{salimans2017evolution,wierstra2014natural} (ES) and Augmented Random Search~\cite{mania2018simple} (ARS) aim to approximate the gradient using finite-differences, and to perform exploration directly in the parameter space by injecting noise, as opposed to exploring in the action space. Later, ~\citet{fortunato2017noisy} found that introducing noise directly to the parameters greatly increased performance when compared to heuristics such as $\epsilon$-greedy policy or entropy regularization. However, all the aforementioned methods rely on computing gradients, either directly or by approximating it, which as a result limits exploration to local neighbourhoods. The first work to successfully apply a non-gradient based method, was able to successfully evolve a 4M+ parameter neural network~\cite{such2017deep} using a genetic algorithm strictly based on mutations and obtain competitive results on modern RL benchmarks. Indeed, the authors found that the GA-evolved network was the only policy not to be beaten by pure Random Search (RS) on a subset of Atari 2600 games, hinting that following the gradient may be more harmful than beneficial in some cases. In~\citet{lehman2018safe}, a safe mutation operator is introduced in order to efficiently explore the parameter space using GAs while avoiding to dramatically alter the network functionalities (i.e. behaviour). The authors showed that the safe mutation operator allows to evolve neural networks with up to 100 hidden layers and argue that their method could be readily used to improve upon the deep GA in~\citet{such2017deep}. Finally, they advocate for an homologous and similarly motivated safe crossover operator to explore different directions and regions of the parameter space, in a safe and principled manner. 

\subsection{Neural Network Representations}

In addition to GAs and parameter space exploration, two other areas of research motivated our paper. The first one is concerned with neural network interpretation and finding common learned representations between networks trained on the same data, from different initializations~\cite{raghu2017svcca,morcos2018insights,li2015convergent}. The second is related to fitness landscape exploration such as stochastic weight averaging~\cite{izmailov2018averaging,athiwaratkun2018there}, cyclical gradient descent~\cite{zhang2019cyclical} and finding potential low-error paths between local minima~\cite{garipov2018loss,goodfellow2014qualitatively}. 

In particular, it was found in~\citet{li2015convergent} that networks trained on the same dataset, starting from different initializations, largely learn the same internal representations. Similarly to our work, the authors investigate the relationships between neurons of two networks in order to find a mapping between them, but do not attempt to recombine neural network weights. They do so by computing cross-correlation of neuron pairs coming from different networks (see Section 3.3). They also seek many-to-one mappings by training a $L_1$ penalized regression model using the neurons in one network as input to predict the activation of a single neuron in the other network. However, this method fails to take into account many-to-many relationships (i.e. colinearity in both the input and output neurons) and requires training of an auxiliary network for each neuron, rendering the method computationally prohibitive. In a later work, ~\citet{raghu2017svcca} proposed to use Canonical Correlation Analysis in order to find many-to-many relationships between the neurons of two networks, without having to train any additional regression models. 

Finally, ~\citet{goodfellow2014qualitatively} studied the loss landscape by linearly interpolating between two networks trained on the same dataset, starting from different initializations and evaluated the loss at evenly spaced intervals. They found that the loss dramatically increases when linearly interpolating between two local minima. However, this is because they perform a naive crossover and do not attempt to solve the competing conventions problem by matching the neurons according to their functionalities.

\section{Correlation Analysis of Neurons}

\subsection{Neuron's Activation Vector}

In order to find a mapping between the hidden layers of two feed-forward neural networks (i.e. a correspondence between the neurons of the two layers) trained on the same dataset, we first have to define how to represent a layer. To do so, we define the representation of a neuron by a vector containing the neuron's activation values for a fixed batch of data points. Formally, we have a dataset $X=\{x_1,\ldots, x_n\}$, where $x_i \in \mathbb{R}^{m}$ denotes the $i$\textsuperscript{th} observation. Then, a neuron would output a scalar value for each of the data points, which means that it can be represented as a vector $h=(g(x_1),\ldots,g(x_n))$, where $g(\cdot)$ denotes the neuron's activation function. Now, it is straightforward to extend this representation to that of a hidden layer, since a layer is made of several neurons. A hidden layer $L$ is simply represented by a matrix in $\mathbb{R}^{n \times p}$ where each of its columns is the vector representation of a neuron, and $p$ is the number of neurons in the layer. In this work, we will consider several pairs of hidden layers $L_a \in \mathbb{R}^{n \times p}$ and $L_b \in \mathbb{R}^{n \times q}$ coming from two neural networks $\theta_a$ and $\theta_b$, trained on the same dataset but with different random initializations.

\subsection{Canonical Correlation Analysis on Hidden Layers}

Canonical Correlation Analysis is a multivariate statistical technique which seeks to find maximally correlated linear relationships between two sets of observations, under orthogonality and norm constraints. In the CCA literature, the two sets of observations are often referred to as views. In this work, we apply CCA to pairs of hidden layers $L_a$ and $L_b$, and use the coefficients of the canonical vectors (see Equation (1)), in order to construct a mapping between the neurons of the two layers. Note that for CCA to work, the number of neurons in each layer does not need to be equal and that CCA is invariant to affine transformations. These two properties make it an ideal tool to be applied to neural networks, since we can compare networks with different topologies and layers at different depths. Furthermore, CCA can be applied to the main types of neural architectures: feed-forward, convolutional and recurrent neural networks~\cite{raghu2017svcca,morcos2018insights}. 

In this paper, however, we aim at providing a proof-of-concept and thus focus our efforts on fully-connected, feed-forward neural networks, and leave the application of safe crossover to more types of architectures as future work. Next, we give an overview of the formulation of CCA, and its basic interpretation in the context of our work. A more detailed account of CCA and its modern variants is given in~\citet{uurtio2018tutorial}. 

Let us consider the two views $L_a \in \mathbb{R}^{n \times p}$ and $L_b \in \mathbb{R}^{n \times q}$, which are both standardized with columns having zero mean and unit variance. The row vectors $x_a^i \in \mathbb{R}^p$ and $x_b^i \in \mathbb{R}^q$ denote the $i$\textsuperscript{th} multivariate observation of $L_a$ and $L_b$ respectively, for $i=1,\ldots,n$. Furthermore, the column vectors $h_a^j \in \mathbb{R}^n$, $j=1,\ldots,p$, and $h_b^j \in \mathbb{R}^n$, $j=1,\ldots,q$ denote the vector representation of the $j$\textsuperscript{th} neuron, of $L_a$ and $L_b$ respectively. Formally, CCA seeks to find linear transformations $z_a = L_a w_a$  and $z_b = L_b w_b$, where $w_a \in \mathbb{R}^p$, $z_a \in \mathbb{R}^n$, $w_b \in \mathbb{R}^q$ and $z_b \in \mathbb{R}^n$, such that the correlation (or equivalently, the cosine of the angle) between $z_a$ and $z_b$ is maximized, subject to orthonormality. Mathematically, CCA can be framed as an optimization problem, where the objective is to sequentially find $w_a^k$ and $w_b^k$, that satisfy the following:

\begin{equation}
    \hat{\rho}^k = \max_{w_a^k, w_b^k} \langle z_a^k, z_b^k\rangle = \max_{w_a^k, w_b^k} \langle L_a w_a^k, L_b w_b^k\rangle,
\end{equation}

subject to

\[||z_a^k||_2=1 \hspace{0.2cm} ||z_b^k||_2=1,\]
\[\langle z_a^k, z_a^r\rangle=0 \hspace{0.2cm} \langle z_b^k, z_b^r\rangle=0,\]
\[\forall \hspace{0.1cm} r \neq k \hspace{0.2cm} \textrm{for} \hspace{0.2cm} r,k=1,\ldots,\textrm{min}(p,q),\]
where $\langle \cdot, \cdot \rangle$ denotes the Euclidean inner product. Terminology wise, $w_a^k$ and $w_b^k$ are the $k$\textsuperscript{th} canonical weights or components and the linear transforms $z_a^k$ and $z_b^k$ are the corresponding canonical variates.

One of the ways to solve the CCA optimization problem in (1) is by using Singular Value Decomposition (SVD)~\cite{healy1957rotation}. Let us denote the covariance matrices of $L_a$ and $L_b$ by $C_{a} \in \mathbb{R}^{p \times p}$ and $C_{b} \in \mathbb{R}^{q \times q}$ respectively, and the cross-covariance between $L_a$ and $L_b$ by $C_{a,b} \in \mathbb{R}^{p \times q}$. Then, we find that the canonical directions are given by
\[w_a = C_{a}^{-\frac{1}{2}}U \hspace{0.2cm} \textrm{and} \hspace{0.2cm} w_b = C_{b}^{-\frac{1}{2}}V,\]
where $U$ and $V$ are the matrices corresponding to the sets of orthonormal left and right singular vectors respectively, which are obtained by solving the following SVD 

\begin{equation}
     C_{a}^{-\frac{1}{2}}C_{a,b}C_{b}^{-\frac{1}{2}} = U^{T}SV.
\end{equation}
In the above equation, $S \in \mathbb{R}^{p \times q}$ is the matrix containing the singular values (in its diagonal entries) of the left-hand-side, which correspond to the canonical correlations.
In summary, for the purpose of our paper, CCA outputs a series of pairwise orthogonal singular vectors $u^k, v^k$ from which the corresponding canonical components $w_a^k = C_{a}^{-\frac{1}{2}}u^k$ and $w_b^k = C_{b}^{-\frac{1}{2}}v^k$ can be computed, alongside the canonical correlation $\hat{\rho}^k \in [0, 1]$, with $\hat{\rho}^k<\hat{\rho}^j$ if $k>j$, for $k=1,\ldots,\textrm{min}(p,q)$.

Note that while CCA is well suited to analyse the relationships between two sets of data, it can overfit to spurious correlation between the two views, in particular in under-determined systems or in data containing a large proportion of noisy variables~\cite{raghu2017svcca,bilenko2016pyrcca}. One needs to be wary of this pitfall since in overparametrized neural networks, many neurons are either redundant or capturing noise in the data~\cite{raghu2017svcca}. To tackle the aforementioned problem, we will apply two different techniques:
Singular Vector Canonical Correlation Analysis (SVCCA)~\cite{raghu2017svcca} and $L_2$ regularized CCA (canonical ridge)~\cite{vinod1976canonical}. This means that when we refer to safe crossover using CCA (SC-CCA), there are two variants: SVCCA and ridge CCA. SVCAA first computes the main variance directions (principal components) by performing SVD on the hidden layers $L_a$ and $L_b$, before carrying out CCA on the lower rank representations of $L_a$ and $L_b$. In doing so, the neurons exhibiting low variance are discarded and CCA is less prone to identifying spurious correlations~\cite{raghu2017svcca}. On the other hand, $L_2$ regularized CCA constrains the norms of the canonical vectors $w_a^k$ and $w_b^k$ in Equation (1) as follows
\vspace{-0.1cm}
\[(w_a^k)^T C_{a}w_a^k+\lambda ||w_a^k||_2^2 =1,\]
\vspace{-0.3cm}
\[(w_b^k)^T C_{b}w_b^k+\lambda ||w_b^k||_2^2=1,\]
which as a result relaxes the orthogonality constraint between the canonical directions, in the original CCA formulation. In this paper, we experiment with different numbers of SVD directions (i.e. the number of principal components kept in the low-rank approximations of $L_a$ and $L_b$) as well as with different regularization values for $\lambda$. In the experiments, we show that the results are sensitive to the number of kept variance directions but not to the value of $\lambda$, in SVCCA and ridge CCA respectively. 

\subsection{Pairwise Cross-Correlation of Neurons} 

\begin{figure*}[t]
    \centering
    \includegraphics[width=0.95\linewidth]{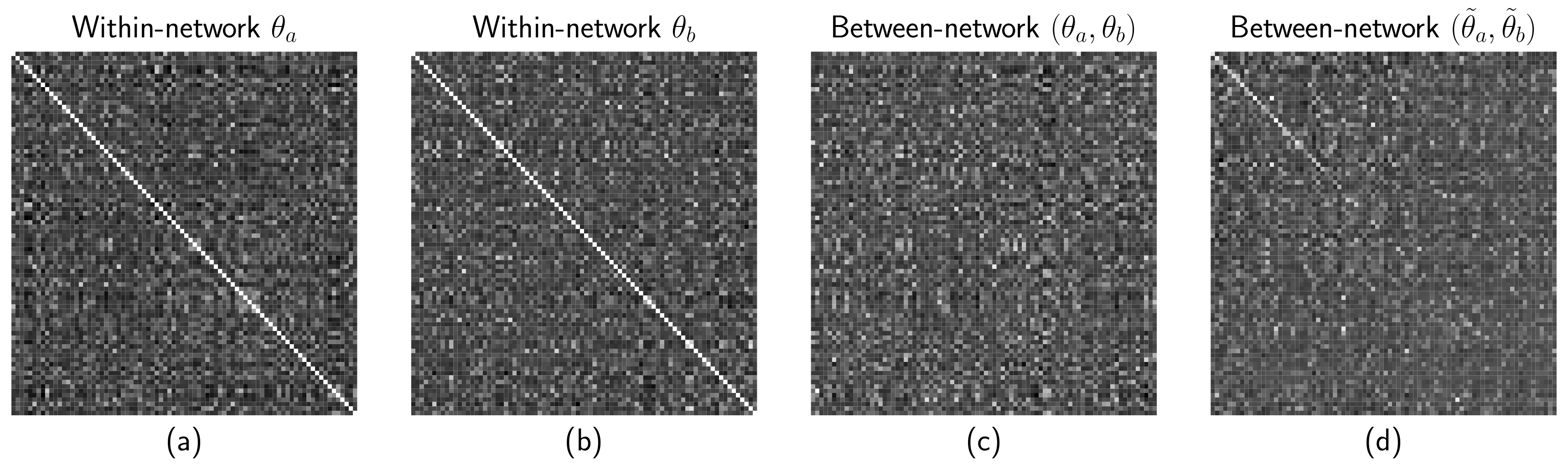}
    \caption{Correlation matrices for neurons of a randomly chosen network pair \boldmath $(\theta_a, \theta_b)$\unboldmath, trained on MNIST. In (a)-(c), we display the pairwise correlation of the first 100 neurons. For (d), we display the correlation of the 100 aligned neuron pairs obtained using SVCCA (keeping 100 variance directions). In other words, the first diagonal element of the matrix in (d) is the correlation between the neuron pair obtained by looking at the coefficients of the first canonical components \boldmath $w_a^1$ \unboldmath and \boldmath $w_b^1$ \unboldmath, and so on.}
    \label{fig:pair_wise_corr}
\end{figure*}

Here, we summarize the approach from~\citet{li2015convergent}, which seeks to find relationships between neuron pairs by computing their cross-correlation. Let $L_a \in \mathbb{R}^{n \times p}$ denote the mean-centred representation of a hidden layer, where each column $h_a^j$, $j=1,\ldots,p$, corresponds to a neuron, as described in Section 3.2. Then, the observed correlation matrix $\widehat{\Sigma}_a \in \mathbb{R}^{p \times p}$ of $L_a$ is defined as

\[\widehat{\Sigma}_{a}^{i,j} = \frac{(h_a^i)^T h_a^j}{\sqrt{\widehat{\textrm{V}}\textrm{ar}(h_a^i)\widehat{\textrm{V}}\textrm{ar}(h_a^j)}}, \hspace{0.2cm} \textrm{for} \hspace{0.2cm} i,j=1,\ldots,p\]
where $\widehat{\textrm{V}}\textrm{ar}(\cdot)$ denotes the observed (sample) variance. Similarly, the observed cross-correlation matrix $\widehat{\Sigma}_{a,b} \in \mathbb{R}^{p \times p}$ between two layers $L_a$ and $L_b$ is defined as 
\[\widehat{\Sigma}_{a,b}^{i,j} = \frac{(h_a^i)^T h_b^j}{\sqrt{\widehat{\textrm{V}}\textrm{ar}(h_a^i)\widehat{\textrm{V}}\textrm{ar}(h_b^j)}}, \hspace{0.2cm} \textrm{for} \hspace{0.2cm} i,j=1,\ldots,p.\]
Note that while $\widehat{\Sigma}_a$ is indeed symmetric, $\widehat{\Sigma}_{a,b}$ is not. In Figure \ref{fig:pair_wise_corr} (a) and (b), the within-network correlation matrices $\widehat{\Sigma}_a$ and $\widehat{\Sigma}_b$ are shown for the first 100 neurons of a randomly chosen pair of networks $\theta_a$ and $\theta_b$, trained on MNIST. On the other hand, in (c), the between-network cross-correlation matrix $\widehat{\Sigma}_{a,b}$ is displayed.

\section{Safe Crossover Operator}

We have now seen how to apply CCA and pairwise cross-correlation to the neurons of neural network pairs. In this section, we are going to describe how to match neurons from two networks by separately using CCA (Section 4.1 and Algorithm 1) and pairwise correlation (Section 4.2). We also introduce an efficient way to permute the neurons of neural networks to be functionally aligned (Section 4.3 and Algorithm 2) and ready to be recombined (Section 4.4).

We will assume that we have trained (with some variants of gradient descent) two feed-forward neural networks of depth $D+1$, $\theta_a$ and $\theta_b$, with identical architectures, on the same dataset, but starting from different initializations. Thus, these networks have $D$ hidden layers and $D+1$ weight matrices. We denote the hidden layer representations and weights matrices (including the biases) as $\{L_a^d\}_{d=1}^{D}$, $\{L_b^d\}_{d=1}^{D}$ and $\{W_a^d\}_{d=1}^{D+1}$, $\{W_b^d\}_{d=1}^{D+1}$ for $\theta_a$ and $\theta_b$, respectively. Note that the rows of the weight matrices correspond to the incoming layer (inputs) and the columns to the outgoing layer (outputs). Furthermore, we denote the functionally aligned versions of $\theta_a$ and $\theta_b$ as $\widetilde{\theta}_a$ and $\widetilde{\theta}_b$, respectively. 

\subsection{Neurons Matching via CCA}

Here, we further assume that we have performed CCA on all the pairs of hidden layers $\{(L_a^d, L_b^d)\}_{d=1}^{D}$. Then, for each pair of layers (at each depth), CCA outputs a series of canonical directions $w_a^k$ and $w_b^k$, $k=1,\ldots,\textrm{min}(p,q)$, for which $\textrm{corr}(w_a^k L_a, w_b^k L_b)$ is maximized. The coefficients of the canonical components $w_a^k$ and $w_b^k$ correspond to the strength of the linear relationship between the neurons of the two layers. The relationship between two neurons from $L_a$ and $L_b$ is positive if their coefficients have the same sign, and is negative if their coefficients have different signs. The strength of the relation is given by the absolute value of the coefficients. 

In Algorithm 1, making use of the aforementioned facts, we describe our method to match up neurons, for a pair of hidden layers $L_a$ and $L_b$. For each pair of canonical components $(w_a^k, w_b^k)$, we seek to match the two neurons with the highest positive relationship. In doing so, we match exactly one pair of neurons per canonical component, for $k=1,\ldots,\textrm{min}(p,q)$. Note that in Algorithm 1, we omit the fact that the same neuron may be chosen more than once: the same neuron in $L_a$ can have strong positive relationships with more than one neuron in $L_b$. In our implementation\footnote{https://github.com/pinouche/GECCO$\_$2020}, if the most recently formed $k$\textsuperscript{th} pair of neurons contains at least one neuron which is already part of a previous pair $j$, $j<k$, then we look for the next highest positive relationship, and proceed in this fashion until all the canonical components have been used. Note that matching neuron pairs by giving priority to the first canonical components is justified since we have that the canonical correlations are ranked in decreasing order with $\hat{\rho}^j<\hat{\rho}^k$ if $j>k$, for $j=1,\ldots,\textrm{min}(p,q)$.

Note that there is a significant difference between Algorithm 1 and the direct application of CCA to the hidden layers proposed in~\citet{raghu2017svcca}. Indeed, in our case, we are using the linear relationships extracted by CCA in order to devise a one-to-one mapping between individual neurons of network pairs. In other words, instead of using CCA to measure some overall similarity between network pairs, we use it to functionally align neural networks in order to introduce a safe crossover operator directly in parameter space (see Section 4.4).

\begin{algorithm}[tb]
\label{alg:algorithm1}
\begin{flushleft}
\textbf{Inputs:}
\begin{enumerate}
    \item[i.] Canonical vectors: $\{w_a^k, w_b^k\}_{k=1}^{\textrm{min}(p,q)}$
    \item[ii.] Empty lists: $l_a$, $l_b$
\end{enumerate}
\textbf{Output:}
\begin{enumerate}
    \item[i.] Lists of ordered neuron indices
\end{enumerate}
\end{flushleft}
\begin{algorithmic}[1] 
\FOR{$k=1,\ldots,\textrm{min}(p,q)$}
\STATE $s_{-}^k$ = $\textrm{abs}(\textrm{min}(w_a^k)+\textrm{min}(w_b^k))$
\STATE $s_{+}^k$ = $\textrm{abs}(\textrm{max}(w_a^k)+\textrm{max}(w_b^k))$  
\IF {$s_{+}^k > s_{-}^k$}
\STATE $l_a[k]=\textrm{argmax}(w_a^k)$, \hspace{0.1cm} $l_b[k]=\textrm{argmax}(w_b^k)$
\ELSE
\STATE $l_a[k]=\textrm{argmin}(w_a^k)$, \hspace{0.1cm} $l_b[k]=\textrm{argmin}(w_b^k)$
\ENDIF
\ENDFOR
\STATE \textbf{return} $l_a$, $l_b$ (ordered neuron indices for layers $L_a$ and $L_b$)
\end{algorithmic}
\caption{Finding pairs of neurons for layers $L_a$ and $L_b$}
\end{algorithm}

\subsection{Neurons Matching via Pairwise Cross-Correlation}

In Section 3.3, we have seen how to compute the cross-correlation matrix $\widehat{\Sigma}_{a,b}$ between two layers $L_a$ and $L_b$. We consider two ways in which the information contained in the matrix can be used to match neurons. 

In the first method, referred to as bipartite semi-matching in graph theory~\cite{lawler2001combinatorial}, each neuron in $L_a$ is paired with the neuron in $L_b$ with which it is maximally correlated. Therefore, more than one neuron in $L_a$ may be paired with the same neuron in $L_b$. In other words, all the neurons in $L_a$ are paired but not necessarily all the neurons in $L_b$. As a result, $\widetilde{\theta}_b$ does not have to be functionally equivalent to the original non-permuted network $\theta_b$, since they may not contain the same neurons. Note that semi-matching yields different results, if each neuron in $L_b$ is instead paired with the neuron in $L_a$ with which it is maximally correlated, due to $\widehat{\Sigma}_{a,b}$ not being symmetric. 

The second method, referred to as bipartite matching, seeks to find a one-to-one pairing of neurons such that the sum of pairwise correlation is maximized~\cite{hopcroft1973n}. In this scenario, all the neurons in both $L_a$ and $L_b$ are used. Therefore, we have that $\widetilde{\theta}_a$ and $\widetilde{\theta}_b$ are functionally equivalent to $\theta_a$ and $\theta_b$, respectively. 

After having applied either of those matching techniques, we obtain a mapping of neurons that can be stored in two lists $l_a$ and $l_b$ (see Section 4.3), similarly to when the matching is made via CCA, in Algorithm 1.

\subsection{Neurons Ordering}

\begin{figure}[tb]
    \centering
    \includegraphics[width=0.7\linewidth]{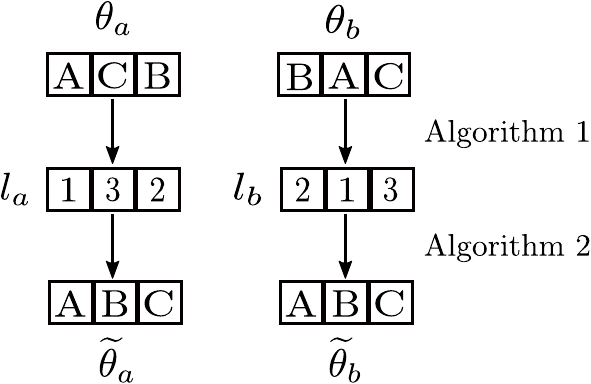}
    \caption{Toy example illustrating how Algorithm 1 and 2 operate. Here, we have two single layered networks \boldmath $\theta_a$ \unboldmath and \boldmath $\theta_b$\unboldmath, which have learned the same features \boldmath $\{A,B,C\}$\unboldmath. Algorithm 1 identifies the pairs of neurons which are functionally identical and returns two lists \boldmath $l_a$ \unboldmath and \boldmath $l_b$ \unboldmath of indices. The following pair of neurons should be formed: (1,2), (3,1) and (2,3). Algorithm 2 uses \boldmath $l_a$ \unboldmath and \boldmath $l_b$ \unboldmath to permute the neurons (i.e. the inbound and outbound weights) and returns the functionally aligned networks \boldmath $\widetilde{\theta}_a$ \unboldmath and \boldmath $\widetilde{\theta}_b$\unboldmath.}
    \label{fig:diagram}
\end{figure}

At each depth $d=1,\ldots,D$, we have two lists $l_a^d$ and $l_b^d$ containing ordered neuron indices. The first element of $l_a$ contains the neuron index from $L_a$ which matches with the neuron index from $L_b$ given in the first element of $l_b$, and so on. Following Algorithm 2, we can then functionally align $\theta_a$ and $\theta_b$ by permuting the neurons of the layers $\{L_a^d, L_b^d\}_{d=1}^{D}$ (and thus the weights) according to the pairings $\{l_a^d, l_b^d\}_{d=1}^{D}$. Finally, once the weights of the two neural networks are permuted, we can safely crossover the two networks by directly matching the weights at the same location in both networks (assuming that the two networks have the same topology). The steps to go from $\theta_a$ and $\theta_b$ to $\widetilde{\theta}_a$ and $\widetilde{\theta}_b$ are illustrated in Figure \ref{fig:diagram}, for a network with a single hidden layer of three neurons. We can view $\widetilde{\theta}_a$ and $\widetilde{\theta}_b$ as non-redundant representations of $\theta_a$ and $\theta_b$, where the networks are now functionally aligned according to a uniquely defined mapping obtained by applying Algorithm 2.

\begin{algorithm}[tb]
\caption{Permuting neural networks weights}
\label{alg:algorithm2}
\begin{flushleft}
\textbf{Inputs:}
\begin{itemize}
    \item[i.] Neuron indices: $\{l_a^d, l_b^d\}_{d=1}^{D}$
    \item[ii.] Neural network weights: $\{W_a^d, W_b^d\}_{d=1}^{D+1}$
\end{itemize}
\textbf{Output:}
\begin{itemize}
    \item[i.] Permuted versions of the input weight matrices
\end{itemize}
\end{flushleft}
\begin{algorithmic}[1] 
\FOR{$d=1,\ldots,D$}
\IF {$d == 1$}
\STATE $\widetilde{W}_a^d = W_a^d[:,l_a^d]$, \hspace{0.1cm}
$\widetilde{W}_b^d = W_b^d[:,l_b^d]$ (order columns)
\ELSE
\STATE $\widetilde{W}_a^d = \widetilde{W}_a^d[:,l_a^d]$, \hspace{0.1cm}
$\widetilde{W}_b^d = \widetilde{W}_b^d[:,l_b^d]$ (order columns)
\ENDIF
\STATE $\widetilde{W}_a^{d+1} = W_a^{d+1}[l_a^d,:]$, \hspace{0.1cm} $\widetilde{W}_b^{d+1} = W_b^{d+1}[l_a^d,:]$ (order rows)
\ENDFOR
\STATE \textbf{return}  $\{\widetilde{W}_a^d\}_{d=1}^{D+1}$ and $\{\widetilde{W}_b^d\}_{d=1}^{D+1}$
\end{algorithmic}
\end{algorithm}

\subsection{Safe Arithmetic Crossover}

We can now use the permuted neural networks $\widetilde{\theta}_a$ and $\widetilde{\theta}_b$ and perform a safe arithmetic crossover on the inbound weights of each neuron (i.e. linearly interpolating the weights), without having to worry about the competing conventions problem and neuron functionalities. In this paper, we consider

\begin{equation}
\widetilde{\theta}_t = (1-t)\widetilde{\theta}_a + t\widetilde{\theta}_b \hspace{0.4cm} t \in [-0.25; 1.25],
\end{equation}
which is simply a weighted average of the two neural networks. Our method could also be used in more complex interpolations (e.g. non-linear curves) and schemes, where only a subset of neurons is recombined at each generation (e.g. symbiotic evolution~\cite{moriarty1996efficient,gomez1999solving}). The naive arithmetic crossover counterpart is simply given by

\begin{equation}
\theta_t = (1-t)\theta_a + t\theta_b, \hspace{0.4cm} t \in [-0.25; 1.25].
\end{equation}

\section{Experiments and Results}

\subsection{Experimental Settings}

In this section, we apply SC-CCA and SC-PwC to feed-forward neural networks trained on MNIST and CIFAR-10. We split the data in the conventional way, using $80\%$ for training and $20\%$ reserved for validation. In this section, the results are reported on the validation sets, which contain 10k examples in both MNIST and CIFAR-10. Again, in this paper, we are much more interested in the relative performance of safe vs naive crossover than in the absolute performance. However, it is of interest to carry out safe crossover on state-of-the-art architectures and more complex datasets in future research. The experimental set-up is the following: 400 unique networks with different random weight initializations are trained (using gradient-descent with Adam) on each dataset, out of which 200 unique pairs are randomly formed. The architecture used on MNIST is a simple network with a single hidden layer of 512 neurons. On CIFAR-10, the network architecture is composed of three hidden layers of 100 neurons each. The activation functions are the same for both datasets: ReLU~\cite{glorot2011deep} on the hidden layers and softmax on the output layer. Weights are initialized using a Normal distribution $N(0,\sqrt{h_i}^{-1})$, where $h_i$ is the number of incoming connections to a given layer from the previous layer’s output (layer $i$). In our experiments, we choose a rather large number of neurons in the hidden layers in order to have a more homogeneous loss distribution, so that the networks learn similar internal representations. Indeed, networks with small hidden layers are more likely to converge to bad local minima~\cite{choromanska2015loss}, implying that they may not learn the same features, rendering any sort of crossover meaningless.

\subsection{Results on MNIST and CIFAR-10}

In this section, the results are presented in two parts: i. the effects of the number of variance directions kept in SVCCA (by first performing SVD on the hidden layers) and of the regulation parameter $\lambda$ in $L_2$ CCA are reported; ii. SC-CCA, SC-PwC and naive crossover are compared on a full (zero-point) arithmetic crossover performed on 200 network pairs, playing the role of parents. Finally, the computational costs of the methods in Section 4.1 and 4.2 are reported.

\subsubsection{SVCCA vs. $L_2$ regularized CCA}

For SVCCA, on CIFAR-10, we chose to keep the same number of variance directions for all three hidden layers. Note that when the number of variance directions is taken to be equal to the number of neurons in the layer (100 and 512 for CIFAR-10 and MNIST respectively), it is equivalent to CCA being performed directly on the hidden layers. In our experiments, on MNIST, we found that keeping 256 of the variance directions lead to better results (the resulting offspring had a lower loss on average) than performing CCA directly on the hidden layers. This is probably due to CCA finding spurious correlations (i.e. finding significant relationships between noisy neurons due to chance), which is more likely to happen for large hidden layers. On the other hand, on CIFAR-10, we found that directly performing CCA on the hidden layer was better than first computing their SVD. For $L_2$ regularized CCA, we try $\lambda \in \{0.01, 0.1, 1\}$ and find that results are consistent across all three values. In particular, for the three $\lambda$ values, at $t=0.5$, we find that the offspring $\widetilde{\theta}_{0.5}$ found with SC-CCA has a lower loss than $\theta_{0.5}$ found using naive crossover, 193 times on MNIST and 200 times on CIFAR-10, out of 200 random and independent trials. In the remainder of this section, for SC-CCA, results are reported using SVCCA with 256 variance directions for MNIST and non-regularized CCA directly applied to the hidden layers for CIFAR-10.

\subsubsection{SC-CCA vs. SC-PwC vs. Naive Crossover} 
\begin{figure*}

\subfloat[]{
	\includegraphics[width=0.45\linewidth]{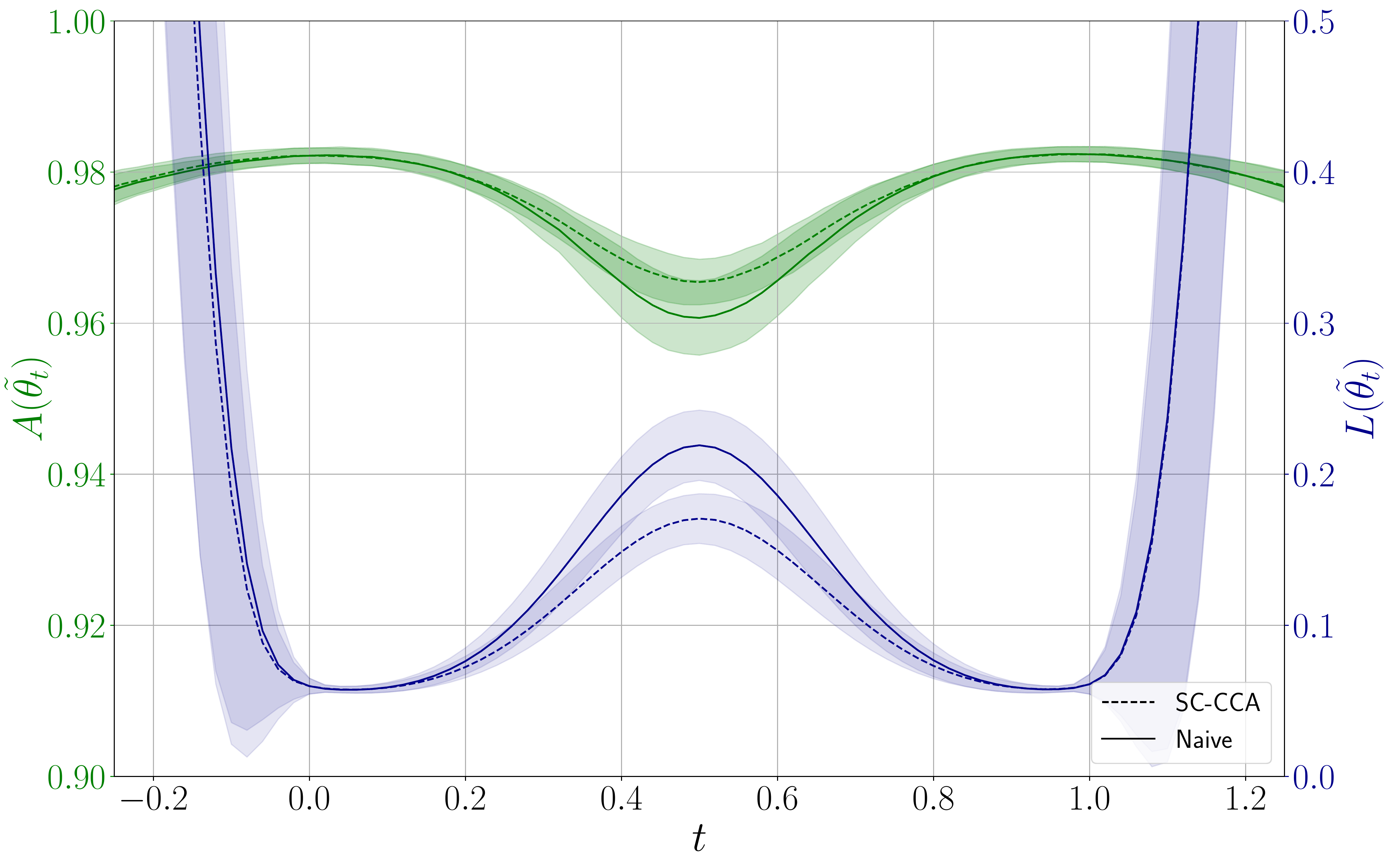} } 
\subfloat[]{
	\includegraphics[width=0.45\linewidth]{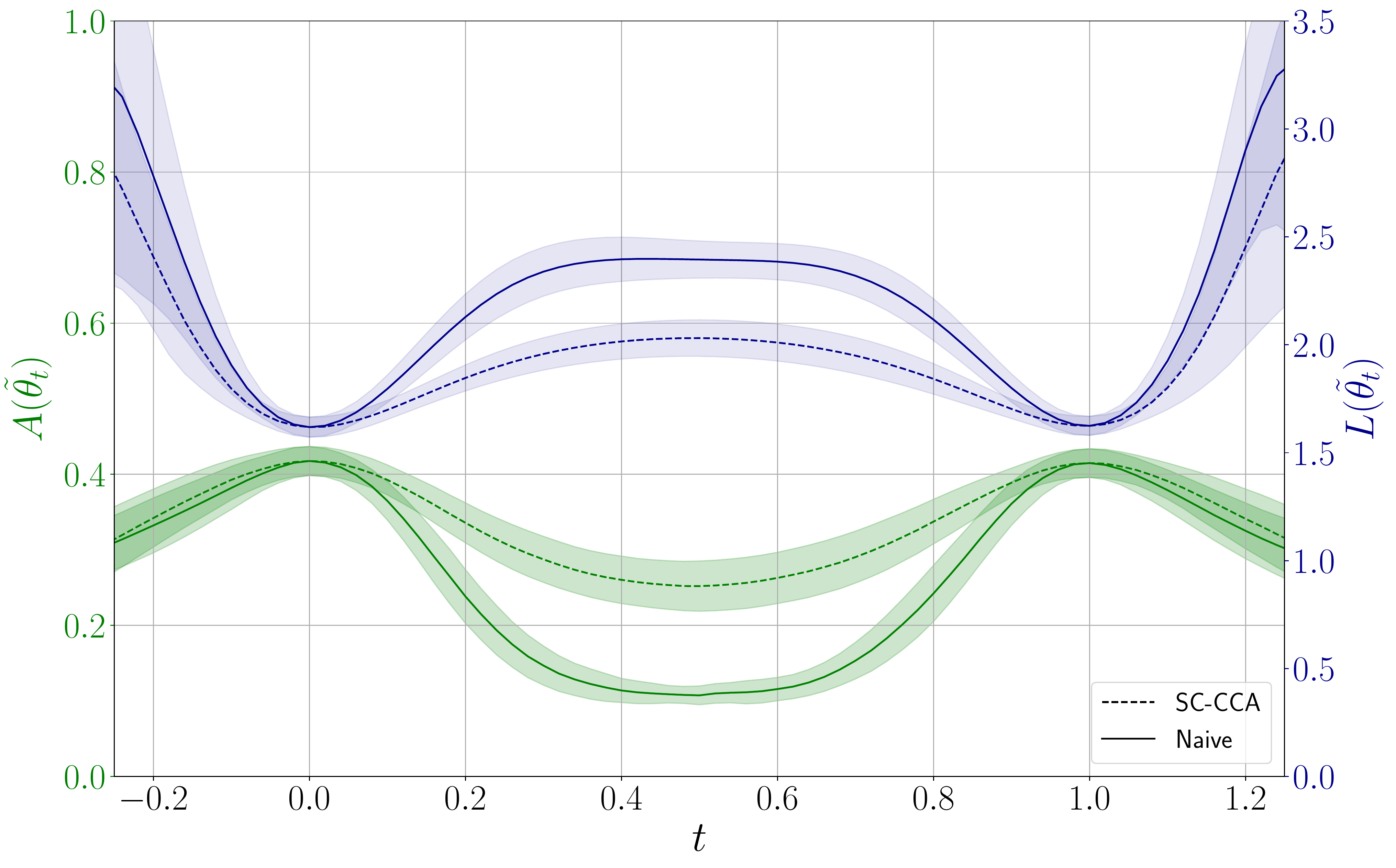} } 
	
\subfloat[]{
	\includegraphics[width=0.45\linewidth]{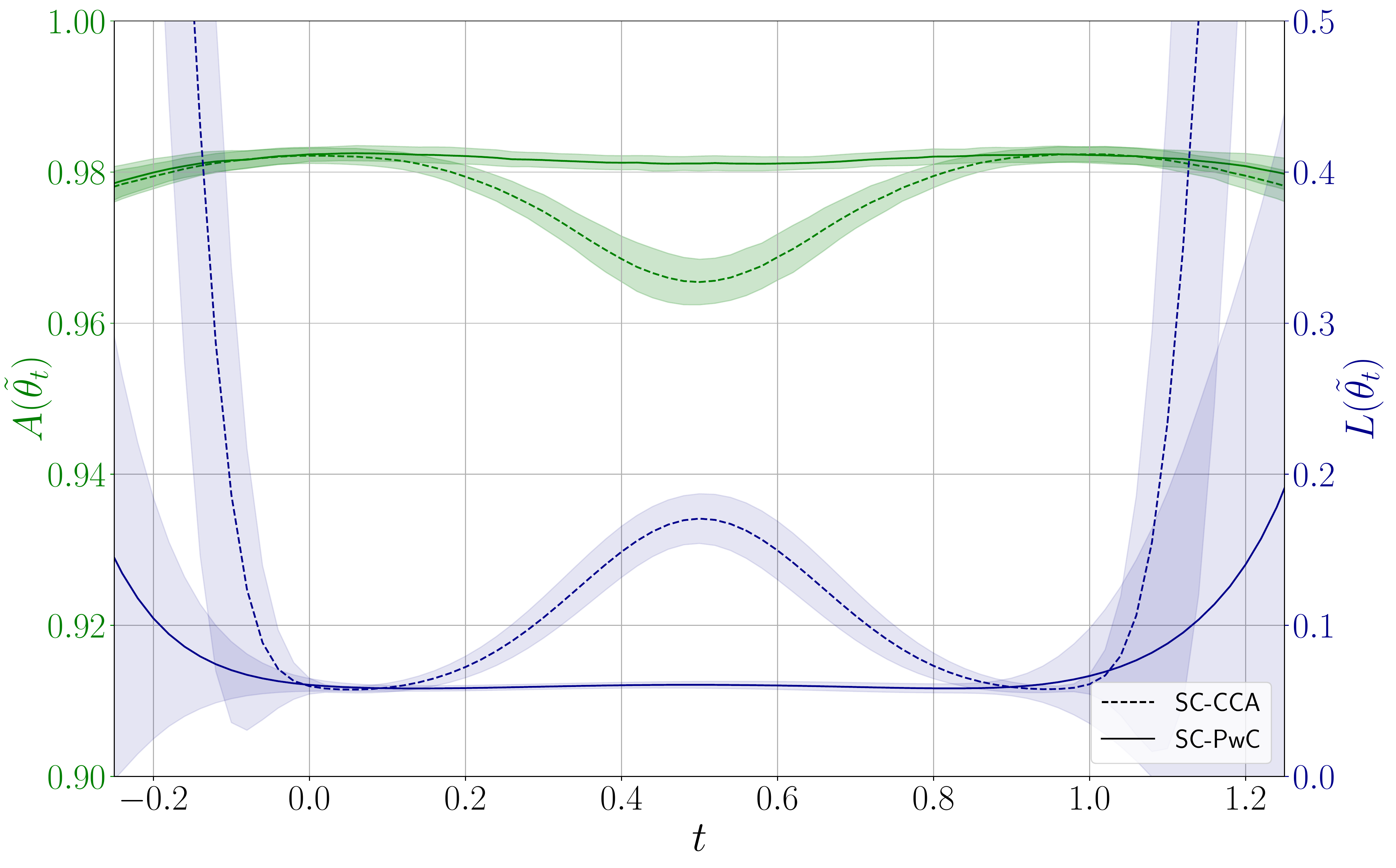} } 
\subfloat[]{
	\includegraphics[width=0.45\linewidth]{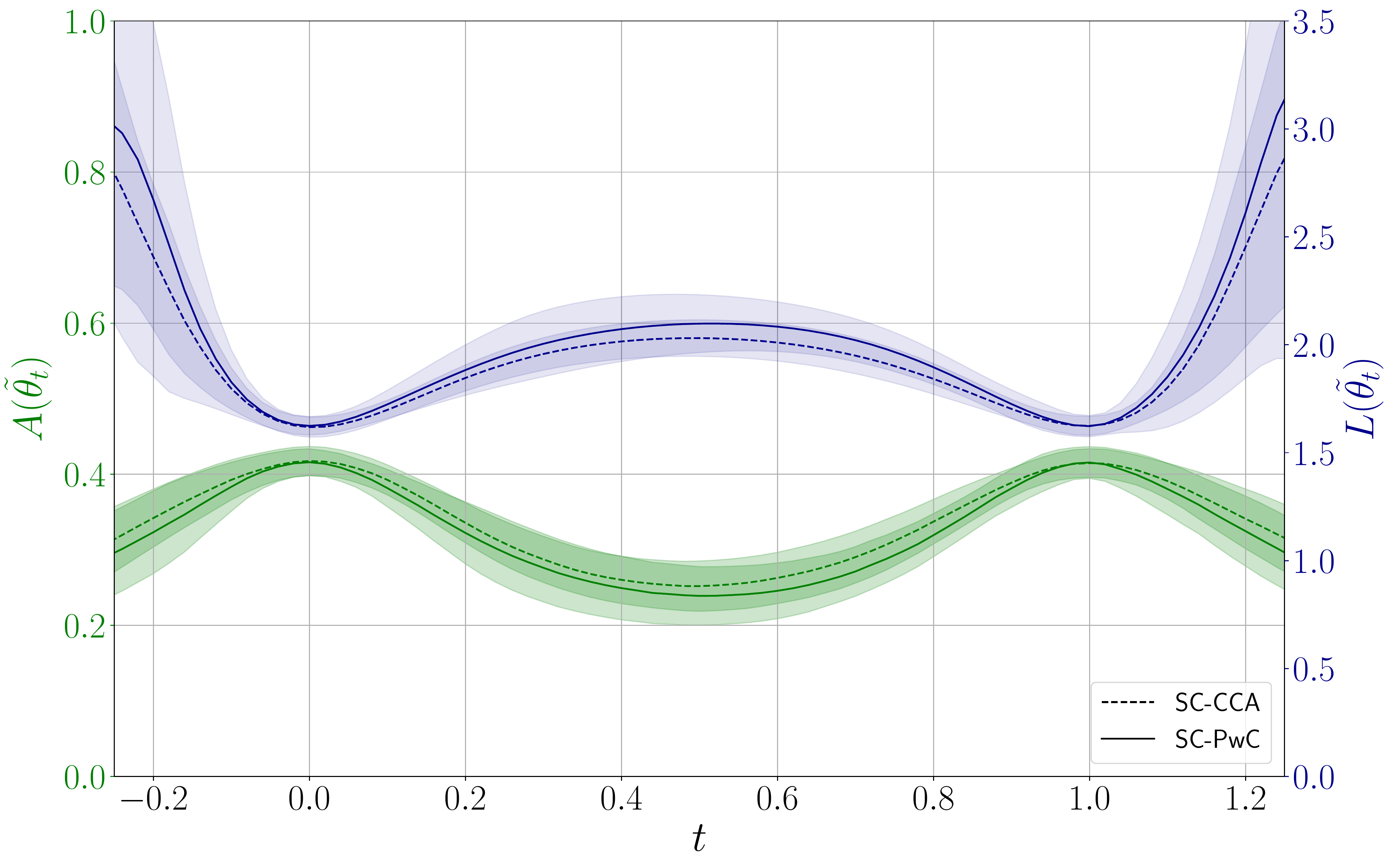} } 
	
\subfloat[]{
	\includegraphics[width=0.45\linewidth]{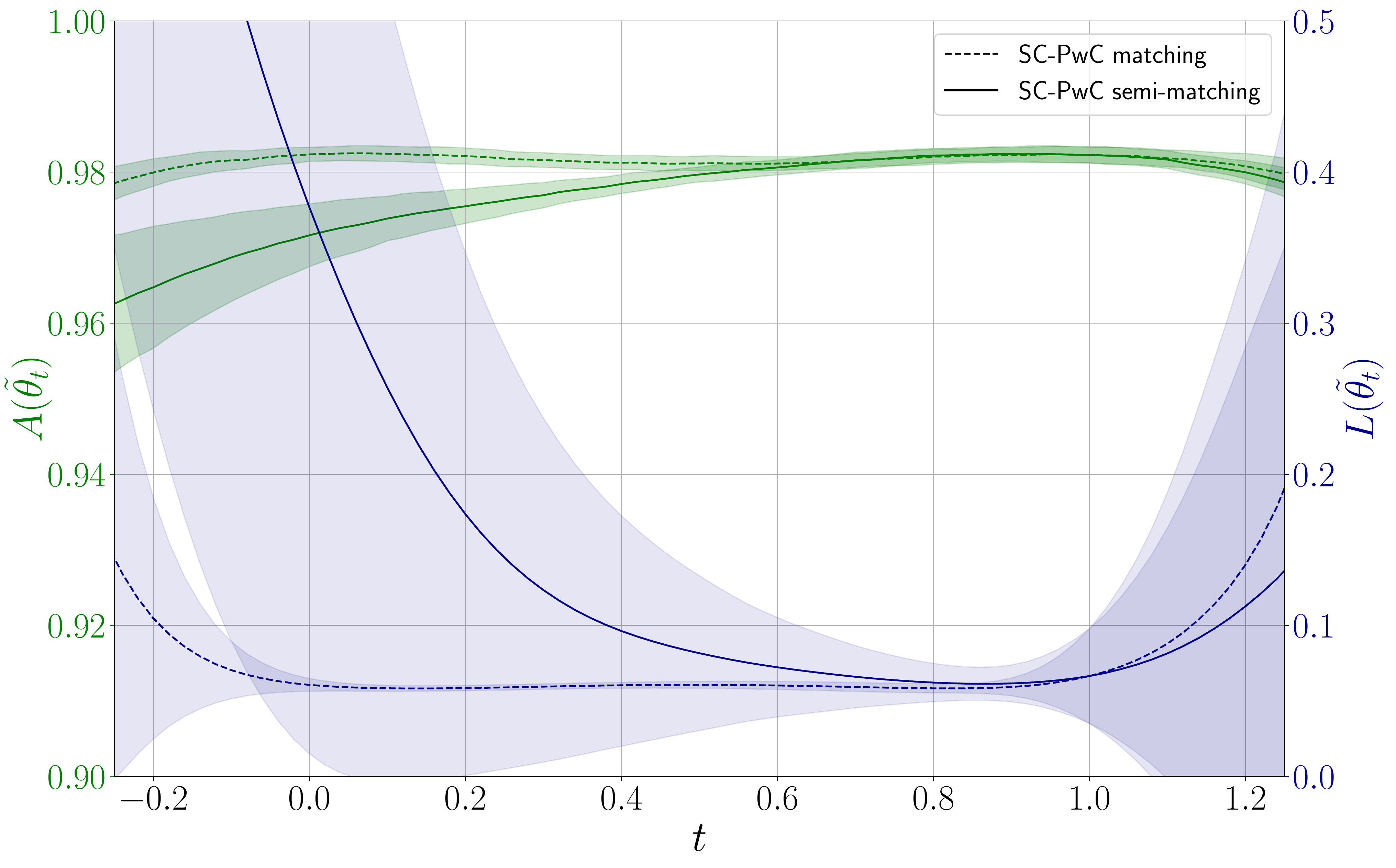} } 
\subfloat[]{
	\includegraphics[width=0.45\linewidth]{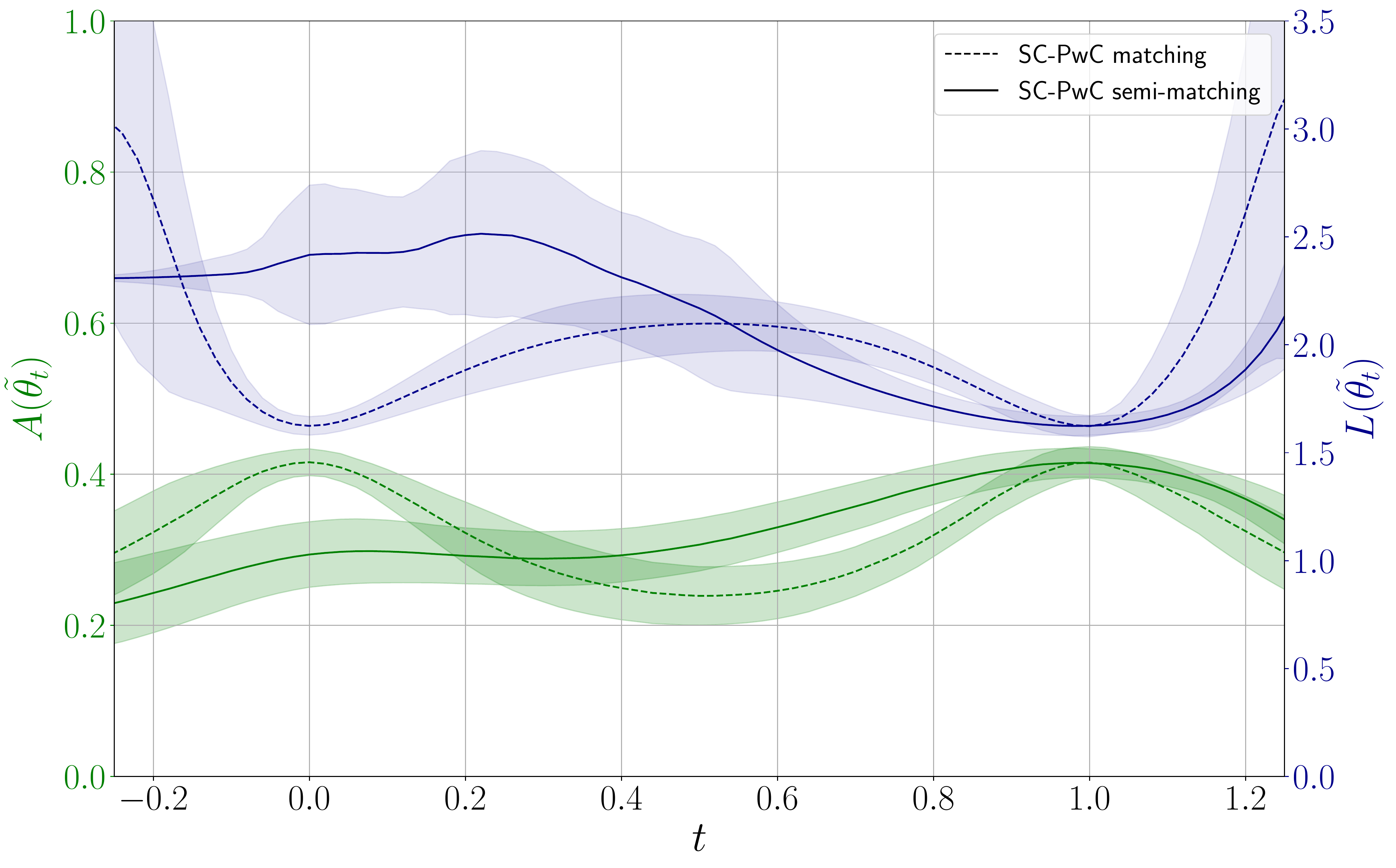} } 

\caption{Comparison of SC-CCA, SC-PwC and naive crossover on MNIST (a), (c), (e) and CIFAR-10 (b), (d), (f). The lines show the average over 200 unique neural network pairs and the shaded areas represent the corresponding standard deviation, computed at regular intervals for \boldmath $t \in [-0.25; 1.25]$ \unboldmath. The left-axis \boldmath $A(\widetilde{\theta}_t)$ \unboldmath in green and the right-axis \boldmath $L(\widetilde{\theta}_t)$ \unboldmath in blue, correspond to accuracy and cross-entropy loss, respectively. On MNIST, in (a), we can see that the gains of SC-CCA are marginal compared to naive crossover, while in (c), SC-PwC significantly improves upon SC-CCA (and thus naive crossover) by consistently finding a very low error path between \boldmath $\widetilde{\theta}_a$ \unboldmath at \boldmath $t=0$ \unboldmath and \boldmath $\widetilde{\theta}_b$ \unboldmath at \boldmath $t=1$ \unboldmath. On the other hand, on CIFAR-10, in (b) and (d), we can see that SC-CCA significantly outperforms naive crossover and also improves upon SC-PwC. 
Indeed, when using naive crossover (b), the accuracy drops all the way down to \boldmath $10\%$ \unboldmath at \boldmath $t=0.5$\unboldmath, which means that our solution becomes as good as random guessing, after having recombined two able parents. Safe crossover makes the accuracy drop to \boldmath $25\%$ \unboldmath, which suggests that the offspring explores the parameter space while retaining some of the parents' functionalities. Finally, in (e) and (f), we compare SC-PwC where neurons are matched using semi-matching or one-to-one matching (see Section 4.2).}
\label{fig:main}
\end{figure*}

Figure \ref{fig:main} shows the results obtained on MNIST and CIFAR-10 when comparing safe (SC-CCA, SC-PwC) and naive crossover. The main takeaway is that safe crossover significantly improves upon naive crossover. Indeed, on CIFAR-10, in (b), when two able parents ($\approx 40\%$ accuracy) are naively recombined, the created offspring drops to $10\%$ accuracy, which is as good as random guessing. On MNIST, in (c), SC-PwC finds a very low-error path between the gradient trained networks $\widetilde{\theta}_a$ and $\widetilde{\theta}_b$. This suggests that for datasets as simple as MNIST, there exists a one-to-one correspondence between neurons. Furthermore, it shows that for a dataset where the classes are well structured and easy to classify, the loss surface between two local minima is approximately flat. However, this is not the case on CIFAR-10, where the loss goes up in-between local minima. It remains to be seen whether this is still the case when more appropriate models such as convolutional networks are used.

Another worthy observation is that while SC-PwC is much better than SC-CCA on MNIST, it produces worse offspring on CIFAR-10. This may be because MNIST is so simple that neurons only learn to recognize a single digit. This makes pairwise cross-correlation an ideal method to match neurons. On the other hand, CCA (much like principal components) extracts more general patterns or "concepts"~\cite{uurtio2018tutorial}, implying that we only match one pair of neurons per concept. In fact, on MNIST, we find that on average (across 200 network pairs) CCA extracts 125 statistically significant relationships per network pair, using Bartlett's test~\cite{bartlett1941statistical}. Meanwhile, with cross-correlation, we find an average of 203 pair of neurons with a correlation of 0.7 or higher, per network pair. The fact that CCA directions are distributed across several neurons~\cite{raghu2017svcca} makes it better suited to capture more complex (many-to-many) relationships between neurons, which naturally occur in more complex datasets. In future work, it is of interest to further investigate the aforementioned observation by comparing the performance of SC-PwC and SC-CCA on datasets such as CIFAR-100 and ImageNet.

\subsubsection{Computational Costs}

Even though SVCCA is a two-step process (first performing the SVD of $L_a$ and $L_b$ and then CCA on the low-rank approximations), it is generally less computationally expensive in practice than running CCA directly on $L_a$ and $L_b$. Indeed, if we assume without loss of generality that $p=q$, then the SVD of $L_a \in \mathbb{R}^{n \times p}$ and $L_b \in \mathbb{R}^{n \times p}$ has complexity of $O(2np^2+2p^3)$ and the complexity of CCA is $O(3np^2+p^3)$. Now, if we denote the low rank approximations of $L_a$ and $L_b$ as $L_a^{1:k} \in \mathbb{R}^{n \times k}$ and $L_b^{1:k} \in \mathbb{R}^{n \times k}$ respectively, then the overall cost of SVCCA is $O(2np^2+2p^3+3nk^2+k^3)$. We are thus comparing $np^2$ to $p^3+3nk^2+k^3$, which is a lot more costly for large $n$ and small $k$ (e.g. $k\leq \frac{p}{2}$). Once we have performed CCA on the hidden layers, the matching method proposed in Algorithm 1 requires only one pass over the canonical vectors and can be solved in $O(p^2)$ time.

On the other hand, the computational cost of matching neurons using pairwise cross-correlation is simply $O(np^2 + p^3)$. The first term $np^2$ corresponds to the computation of the cross-correlation matrix between $L_a$ and $L_b$ and the second term corresponds to the cost of using bipartite matching~\cite{hopcroft1973n} in order to maximize the sum of pairwise correlations.

The computational cost of these two methods is thus similar and is negligible when compared to the computational cost required to train these networks in the first place. Indeed, for a network with $p_w$ learnable parameters (i.e. weights and biases), the cost of one epoch (using all the data points once) is $O(Np_w)$, with $N$ usually much larger than $n$ used in computing CCA or pairwise cross-correlation.

\section{Conclusion and Future Work}
In this paper, we proposed two safe crossover operators acting directly on neural network parameters. We showed, on MNIST and CIFAR-10, that they significantly outperformed naive crossover when linearly interpolating between two neural networks trained on the same dataset, from different initializations. This new operator allows us to explore different regions of the parameter space, without erasing the internal representations learned by the parents. Furthermore, mapping functionally equivalent networks to a unique non-redundant representation greatly reduces the size of the search space, which is particularly relevant when using global optimization algorithms. 

In future research, it is of interest to apply safe crossover to state-of-the art architectures and more complex datasets such as ImageNet. This work could also serve as the basis to develop more methods to perform safe crossover in parameter space of neural networks. In addition, metrics to compare different safe crossover operators according to how much exploration (i.e. how different the offspring are to the parents) is performed relative to how much information is lost could be developed and investigated. 

More related to genetic algorithms, a straightforward research line would be to use safe crossovers in GAs to evolve neural network weights, in RL tasks~\cite{such2017deep}. The hope is that safe crossover would improve the efficiency of GAs in a similar way that they benefited from safe mutations~\cite{lehman2018safe}. 

Furthermore, safe crossover could enable training of distributed data-parallel neural networks. Instead of training a single network, simultaneously evaluating the gradient on several batches and averaging gradients for the backward pass, one could train several networks in parallel and average their parameters. This can be seen as a form of networks ensemble where the ensembling happens in parameter space~\citet{izmailov2018averaging}, as opposed to output space. Finally, instead of using a traditional model ensemble (e.g. the arithmetic crossover used in this paper) or model stacking, we could use more advanced evolutionary algorithms such as symbiotic evolution~\cite{moriarty1996efficient,gomez1999solving}, where only a subset of neurons is recombined (ensembled) at each generation.

\newpage


\end{document}